\pdfoutput=1

\documentclass{article}

\PassOptionsToPackage{numbers, compress}{natbib}
\usepackage[final]{nips_2016}
\usepackage[utf8]{inputenc} % allow utf-8 input
\usepackage[T1]{fontenc}    % use 8-bit T1 fonts
\usepackage{url}            % simple URL typesetting
\usepackage{booktabs}       % professional-quality tables
\usepackage{amsfonts}       % blackboard math symbols
\usepackage{nicefrac}       % compact symbols for 1/2, etc.
\usepackage{microtype}      % microtypography
\usepackage{longtable}
\usepackage{graphicx}
\graphicspath{{./figures/}}
\DeclareGraphicsExtensions{.png}

\usepackage{amsmath}
\usepackage{algorithm}
\usepackage{subcaption}
\usepackage{natbib}
\newcommand{\ie}{i.e.}%{\emph{i.e.}}
\newcommand{\eg}{e.g.}%{\emph{e.g.}}
\newcommand{\vs}{vs}%{\emph{vs}}
\newcommand{\etal}{et al.}%{\emph{et al.}}
\newcommand{\fig}[1]{Fig.~\ref{fig:#1}}
\newcommand{\sect}[1]{Sect.~\ref{sec:#1}}
\newcommand{\tab}[1]{Table ~\ref{tab:#1}}
\newcommand{\equ}[1]{Eq.~\eqref{eq:#1}}
\newcommand{\alg}[1]{Algorithm~\ref{alg:#1}}

\DeclareMathOperator*{\argmin}{\arg\min}
\DeclareMathOperator*{\argmax}{\arg\max}
\newcommand{\sour}{s}
\newcommand{\tar}{t}
\newcommand{\adapt}{{t_a}}
\title{Node-Adapt, Path-Adapt and Tree-Adapt: Model-Transfer Domain Adaptation for Random Forest}

\author{
  Azadeh S.~Mozafari\\
  Department of Computer Engineering\\
  Sharif University of Technology\\
  Tehran, Iran\\
  \texttt{amozaffari@ce.sharif.edu} \\
  \And
  David Vazquez\\
  Computer Vision Center\\
  Universitat Autonoma de Barcelona\\
  Barcelona, Spain\\
  \texttt{dvazquez@cvc.uab.es}\\
  \And
  Mansour Jamzad\\
  Department of Computer Engineering\\
  Sharif University of Technology\\
  Tehran, Iran\\
  \texttt{jamzad@sharif.edu} \\  
  \And
  Antonio M.~Lopez\\
  Computer Vision Center\\
  Universitat Autonoma de Barcelona\\
  Barcelona, Spain\\
  \texttt{antonio@cvc.uab.es}\\
}

\begin{document}
% \nipsfinalcopy is no longer used

\maketitle

\begin{abstract}
Random Forest (RF) is a successful paradigm for learning classifiers due to its ability to learn from large feature spaces and seamlessly integrate multi-class classification, as well as the achieved accuracy and processing efficiency. However, as many other classifiers, RF requires domain adaptation (DA) provided that there is a mismatch between the training (source) and testing (target) domains which provokes classification degradation. Consequently, different RF-DA methods have been proposed, which not only require target-domain samples but revisiting the source-domain ones, too. As novelty, we propose three inherently different methods (Node-Adapt, Path-Adapt and Tree-Adapt) that only require the learned source-domain RF and a relatively few target-domain samples for DA, {\ie} source-domain samples do not need to be available. To assess the performance of our proposals we focus on image-based object detection, using the \emph{pedestrian detection} problem as challenging proof-of-concept. Moreover, we use the RF with \emph{expert} nodes because it is a competitive \emph{patch-based} pedestrian model. We test our Node-, Path- and Tree-Adapt methods in standard benchmarks, showing that DA is largely achieved.     
\end{abstract}

\section{Introduction}
Random Forest (RF) is a paradigm that allows to learn an ensemble of randomized decision-tree classifiers \cite{Breiman:2001, Criminisi:2012}. RF is very successful due to its ability to learn from large feature spaces and seamlessly integrate multi-class classification, as well as the accuracy and processing efficiency achieved by the ensemble. Note that the evaluation of the trees can be parallelized (their training too) or a sequential soft-cascade used \cite{Bourdev:2005}; moreover, evaluating a relatively well-balanced tree shows logarithmic complexity with the number of nodes. We can find recent paradigmatic applications such as its use for pose recognition in depth images (Kinect) \cite{Shotton:2011}, or as classification layer in the GoogLeNet deep CNN architecture for large-scale image classification \cite{Kontschieder:2015}. 

Usually, when learning any classifier, including RF, it is assumed that the training and testing samples follow the same probability distribution. However, in practice, there are application-dependent reasons that make this assumption to fail. The consequence is that the classifier performs worse than expected. In this case, we have to augment the machine learning formalism with the concepts of \emph{source} and \emph{target} domains. The training samples are acquired in the former, and the testing ones in the latter. Then, the challenge is to define a domain adaptation (DA) method which allows to adapt the source-domain-trained classifier for performing well in the target domain, with a minimum effort. By \emph{minimum effort} we refer to several assumptions. For instance, it is assumed little data labeling effort, {\ie} either we have relatively few labeled target-domain samples to perform the (\emph{supervised}) DA, or we have eventually many target-domain samples but without labels (\emph{unsupervised} DA). Moreover, DA methods that use the source-domain classifier and target-domain samples (\emph{model-transfer}) are preferred over those relying on the source-domain samples (\emph{feature-transfer}). Note that in some cases it may not be possible to have access to the source-domain samples ({\eg} a trained classifier can be available for public use, while the data used for its training may be confidential), and in any case the DA used to be slower and more memory demanding. 

In this paper, we focus on supervised model-transfer DA for RF classifiers. In fact, as we will summarize in \sect{relatedWork}, there are previous DA methods for RF; however, they require revisiting the source-domain samples. Thus, aiming model-transfer DA for RF is a novelty of this paper. Actually, after quickly reviewing the specific RF paradigm used here (\sect{RF}), we will propose  three inherently different methods (\sect{RFDA}), termed as \emph{Node-Adapt}, \emph{Path-Adapt} and \emph{Tree-Adapt}, according to the level at which adaptation is performed.

Model-tranfer DA is especially relevant when dealing with complex data such as images. Therefore, in this paper, to assess the performance of our proposals, we focus on image-based object detection, using the \emph{pedestrian detection} problem as challenging proof-of-concept. Moreover, we use the RF with \emph{local expert} splitting nodes (RF-LE) \cite{Yao:2011, Marin:2013, Ristin:2016} because it allows to learn a \emph{patch-based} pedestrian model which is competitive even with more elaborated deep CNNs \cite{Fukui:2015}. As we will see (\sect{experiments}), we test our Node-, Path- and Tree-Adapt methods in standard benchmarks, showing that DA is largely achieved. Interestingly, we will show how a pedestrian detector based on a RF-LE trained with synthetic data (no manual labeling required) together with a relatively low number of real-world labels, performs not too far than an equivalent detector based on the labeling of 10-20 times more pedestrians. Finally, \sect{conclusions} draws our conclusions.
   
\section{Related works}
\label{sec:relatedWork}

Several methods have been proposed to face the DA problem for RF. Goussies {\etal} \cite{Goussies:2014} consider supervised transfer learning for gesture and character recognition. They propose a method for tuning the splitting function parameters of each node, so that the information gain of the target and source-domain samples decrease simultaneously. Danielsson and Aghazadeh \cite{Danielsson:2014} focus on unsupervised DA to adapt a pose estimator learned in synthetic data to operate in real scenarios. They propose an optimization function which selects the best parameters for the splitting function in each node. This is done by minimizing the information gain of the source-domain samples while minimizing the chi-square distance between the source and target feature histograms. Vezhnevets and Buhmann \cite{Vezhnevets:2011} performed DA for the so-called \emph{semantic texton forest} used for semantic segmentation. They consider the same theoretical setting as \cite{Danielsson:2014} but using L1-norm for calculating the distance between two histograms instead of the chi-square. 

All these RF-DA methods require the source-domain samples, {\ie} they follow the feature-transfer paradigm. In contrast, model-transfer DA methods are eventually more interesting since the source data is not always available and they tend to be more computationally efficient; thus, opening the door to hierarchical and online adaptation strategies. In fact, there are already model-transfer DA proposal for popular successful classifiers such as SVM \cite{Yang:2007, Aytar:2011, Mozafari:2016}, structural SVM (SSVM) \cite{Xu:2014b}, deep CNNs \cite{Hoffman:2014}, SSVM in hierarchical target domains \cite{Xu:2016}, and SSVM in an online setting \cite{Xu:2016b}. In this paper, as novelty, we aim at proposing and evaluating three complementary model-transfer DA methods for the RF-LE.

\section{Random forest of local experts}
\label{sec:RF}

Random Forest is a paradigm that allows to learn an ensemble of randomized decision-tree classifiers from a set of training samples. The classification of a testing sample depends on the average classification probabilities over all the trees. Randomness is the core to produce non-correlated decision trees which together lead to a highly accurate ensemble. In this paper, we focus on the RF-LE proposed in \cite{Marin:2013} for binary classification ({\ie} \emph{pedestrian} {\vs} \emph{background}), which is competitive for pedestrian detection \cite{Fukui:2015}. In this case the training samples come in the form of image windows with a canonical size and the label that indicates if the window contains either a pedestrian or background. Randomness is introduced at splitting nodes. In particular, many possible rectangular sub-windows within the canonical window are considered (\emph{patches}), but just a randomly chosen subset is selected. The samples reaching a splitting node are evaluated according to such selected patches. For each patch a combination of its HOG and LBP features is feed into a linear SVM learning process. The patch that allows to learn the \emph{best} linear SVM is chosen as the \emph{local expert} that remains attached to the splitting node. Here, \emph{best} is defined in terms of the purity of the two sample partitions (considering only the samples reaching the node) induced by applying each linear SVM with an optimum score threshold. In the following we summarize this procedure more formally.  

Let $S_j$ be the set of samples reaching the splitting $j^{th}$ node of the tree under training. Let $h(\vec{v};\theta_j)\in\{l,r\}$ be the splitting parametric function at node $j^{th}$, being $\theta_j$ the set of parameters and $\vec{v}\in S_j$ a feature vector (representing a sample). The optimal $\theta_j$ is learned as $\theta_j = \argmax_{\theta} I(S_j;\theta)$, where $I(S_j;\theta)$ is the information gain defined by:
\begin{equation} \label{eq:informationGain}
I(S_j;\theta) = H(S_j)- ((|S_{j,\theta}^l|/|S_j|)H(S_{j,\theta}^l)+(|S_{j,\theta}^r|/|S_j|)H(S_{j,\theta}^r)), H(S) \equiv \mbox{\scriptsize\emph{ClassEntropyOf}}(S),\\
\\
\end{equation}
\noindent where $\theta$ induces a two-subset split of $S_j$ ($S_j = S_{j,\theta}^l \cup S_{j,\theta}^r$) by evaluating $S_j$ with $h(\vec{v};\theta)$; {\ie} $\forall \vec{v}\in S_j$, $\vec{v}$ will go to $S_{j,\theta}^c$ if $h(\vec{v};\theta)=c, c\in\{l,r\}$. More precisely, $\theta = (\phi,\vec{\psi},\tau)$, so that 
$h(\vec{v};\theta)=\{\vec{\psi}\cdot\phi(\vec{v})+\tau\geq0: l; r\}$. Following \cite{Marin:2013}, $\vec{v}$ contains the HOG+LBP features representing a given canonical window ({\ie} a sample), $\phi(\vec{v})$ contains the HOG+LBP features of a selected sub-window within the canonical window, and $\vec{\psi}\cdot\phi(\vec{v})$ consists in applying a linear SVM ($\vec{\psi}$ is the vector of weights, and "$\cdot$" means scalar product). All $\theta_j$ are fixed at testing time, at training time randomness is introduced while looking for the best $\vec{\psi}$ at each splitting node. \alg{trainNode} summarizes the learning procedure for any splitting node. A node becomes a leaf node if either a maximum tree depth is reached, or the number of samples reaching the node is below a threshold (no sense to split them more), or if the percentage of samples belonging to a class is above a threshold. The percentage of positive and negative samples ({\eg} pedestrians and backgrounds) reaching a leaf node determines the posterior probability distribution attached to the node, {\ie} the node $P(y|\vec{v})$, being $y$ the class label (\emph{pedestrian} or \emph{background}). The different trees of the forest are trained independently. 

A sample $\vec{v}$ under classification is processed by all the forest' trees (but soft-cascade can be used for speeding up the process \cite{Marin:2013}), which is done by thresholding the following posterior probability:
\begin{equation} \label{eq:rf-ppd}
P(y|\vec{v})=\frac{1}{T}\sum_{i=1}^T P_i(y|\vec{v}),
\end{equation}
\noindent where $T$ is the number of trees in the forest, and $P_i(y|\vec{v})$ is the posterior probability that tree $i$ assigns to sample $\vec{v}$. For obtaining $P_i(y|\vec{v})$, $\vec{v}$ is routed through the tree $i$ by the splitting nodes until reaching a leaf node, and the $P(y|\vec{v})$ at this leaf node is taken as the final $P_i(y|\vec{v})$ of the tree.  

\begin{algorithm}[t] 
\caption{Training steps for the $j^{th}$ node of the Random Forest of Local Experts (details in \cite{Marin:2013})}
\label{alg:trainNode}
\begin{description}
\item[1)] $K$ feature selectors (patches) {$\phi_1(\vec{v}),...,\phi_K(\vec{v})$} are selected randomly.
\item[2)] For $k=1,...,K$ do:
\begin{description}
\item[2.1)] Let $S_{j,\phi_k}$ be the transformed set of samples, $S_{j,\phi_k} = \left\{\phi_k(\vec{v}):\vec{v}\in S_j\right\}$.
\item[2.2)] Learn the linear SVM weights $\vec{\psi}_k$ with $S_{j,\phi_k}$.
\item[2.3)] Let $\theta=(\phi_k,\vec{\psi}_k,\tau)$ and $h(\vec{v};\theta)=\{\vec{\psi}_k\cdot\phi_k(\vec{v}) + \tau \geq 0: l; r\}$. \\
            Let $\tau_k = \argmax_{\tau} I(S_{j,\phi_k};\theta_k)$ according to \equ{informationGain} and then $\theta_k=(\phi_k,\vec{\psi}_k,\tau_k)$.
\end{description}
\item[3)] Let $j = \argmax_{k\in\{1,\ldots,K\}} I(S_{j,\phi_k};\theta_k)$. Then $\theta_j=(\phi_j,\vec{\psi}_j,\tau_j)$.
\end{description}
\end{algorithm}

\section{Domain adaptation for random forest}
\label{sec:RFDA}

We assume the so-called \emph{covariate shift} commonly considered for object detection problems \cite{Jiang:2008}; {\ie} the marginal distribution of the source- and target-domain samples are different $(P^\sour(\vec{v}^\sour)\neq P^\tar(\vec{v}^\tar))$ while their posterior distributions are similar $(P^\sour(y|\vec{v}^\sour)\simeq P^\tar(y|\vec{v}^\tar))$. Then, in model-transfer DA, we can search for a classifier in the target-domain hypothesis space which can discriminate the available target-domain samples properly, while being similar to the source classifier which acts as regularizer (prior knowledge). This regularization is important since we assume a supervised DA setting where the number of target-domain samples is relatively low. In particular, we have a set $S^\adapt$ of $m$ target-domain labeled samples and a RF-LE ($F^\sour$) trained on $n$ source-domain samples, $m \ll n$. Overall, our final goal is to learn an accurate RF-LE ($F^\adapt$), using $S^\adapt$ and $F^\sour$ as regularizer. To achieve this, we present three novel and inherently different supervised model-transfer DA methods: Node-Adapt (\sect{node-adapt}), Path-Adapt (\sect{path-adapt}), and Tree-Adapt (\sect{tree-adapt}). 

In Node- and Path-Adapt we build a forest $F^\adapt$ only using $S^\adapt$ and forcing it to match as much as possible the structure of $F^\sour$. Briefly, the number of trees is the same ($T$), and the feature selectors at nodes ($\phi_j$) as well as the maximum depth of each path are not changed.  In Node-Adapt such adaptation is done at node level, both the linear SVM weights ($\vec{\psi}_j$) and thresholds ($\tau_j$) are adapted. In Path-Adapt, the adaptation is done simultaneously for all the nodes of each root-to-leaf path, but only the thresholds ($\tau_j$) are adjusted. Tree-Adapt consists in a \emph{reforestation}. A ratio $C$ of randomly selected trees of $F^\sour$ is replaced by the same number of trees trained with $S^\adapt$ from the scratch.

\begin{figure}[t] 
    \begin{subfigure}[b]{0.3\textwidth}
        \includegraphics[height=5cm, width=7cm]{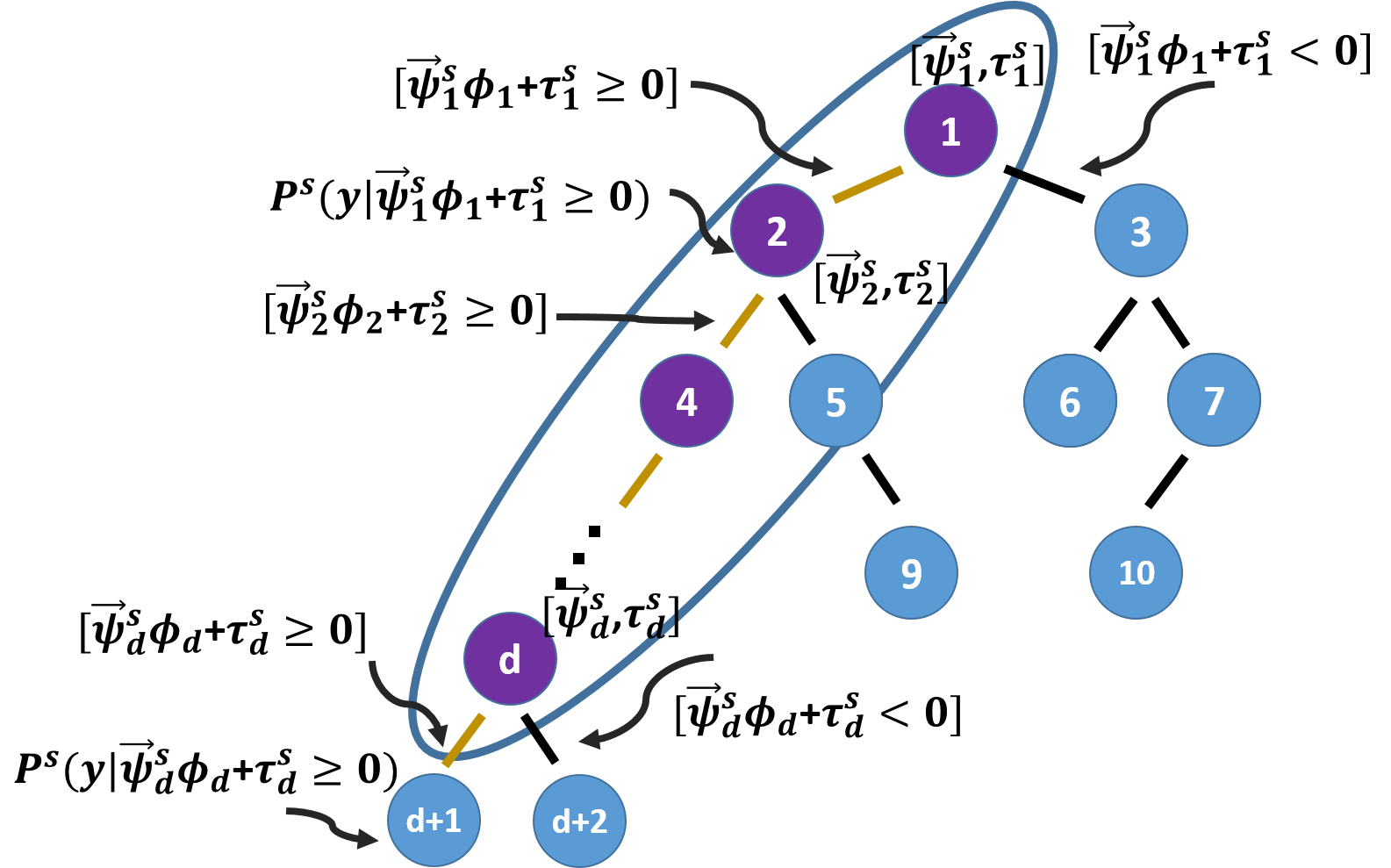}
        \caption{Source-domain tree.}
        \label{fig:source}
    \end{subfigure}\quad\quad\quad\quad\quad\quad\quad\quad\quad    
    \begin{subfigure}[b]{0.3\textwidth}
        \includegraphics[height=5cm, width=7cm]{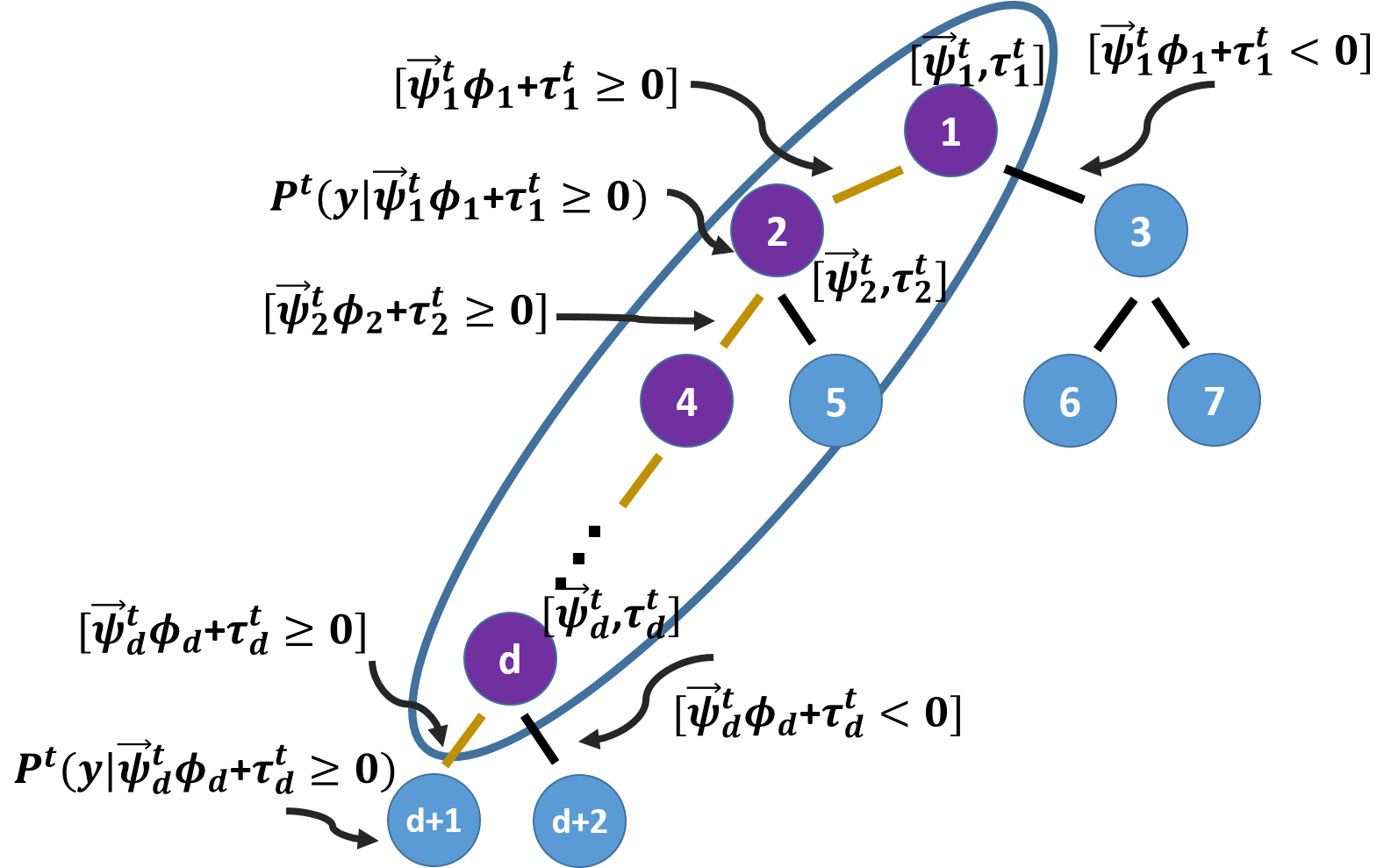}
        \caption{Reshaped target-domain tree.}
        \label{fig:target}
    \end{subfigure}    
    \caption{Node- and Path-Adapt rely on the structure of source trees, which are reshaped during DA. \equ{demoCSNodeAdapt} is derived for the highlighted paths of the corresponding source- and target-domain trees.}\label{fig:cloningTrees}
\end{figure}

\subsection{Node-Adapt}
\label{sec:node-adapt}

Node-Adapt applies a node level adaptation separately for each of the trees cloned from $F^\sour$; thus, we can focus on how this is done for the $i^{th}$ tree by using $S^\adapt$. The method proceeds as follows (see \fig{cloningTrees}). The first node of $i$ to be adapted is the root, whose local expert is parameterized by $\theta_j^\sour=(\phi_j^\sour,\vec{\psi}_j^\sour,\tau_j^\sour)$ for $j=1$, where the upper script reminds that it was learned in the source domain. The adaptation will act on $\vec{\psi}_j^\sour$ and $\tau_j^\sour$ but the feature selector $\phi_j^\sour$ does not change, so we can just denote it as $\phi_j$. In particular, Node-Adapt applies the Adaptive SVM (A-SVM) \cite{Yang:2007}, a model-transfer convex DA method, to obtain an adapted SVM hyper-plane $\vec{\psi}_j^\adapt$ by using $S_j^\adapt$ and $\vec{\psi}_j^\sour$. This is done by solving the following optimization problem formulated for any node $j$ (for the root $j=1 \wedge S_1^\adapt = S^\adapt$):
\begin{equation} \label{eq:a-svm}
\begin{array}{ll}
\vec{\psi}_j^\adapt=\argmin_{\vec{\psi}_j}\frac{1}{2}\|\vec{\psi}_j-C_1\vec{\psi}_j^\sour\|^2+C_2\sum_{k=1}^{|S_j^\adapt|}\epsilon_k ,\\
&\\
s.t.: \quad \epsilon_k\geq 0 \quad \wedge \quad y_k.\vec{\psi}_j\cdot\phi_j(\vec{v}_k^\adapt)\geq 1-\epsilon_k\quad \forall k\in \left\{1,...,|S_j^\adapt|\right\},\\
\end{array}
\end{equation}
\noindent where $y_k$ is the class label of the target-domain feature vector $\vec{v}_k^\adapt$. Note that class labels do not change from source to target domains, so we do not need to distinguish domains for them. The main goal of \equ{a-svm} is to obtain a hyper-plane ($\vec{\psi}_j^\adapt$) which can separate the different classes properly in target domain while, according to covariate shift at node level, being similar to the source-domain hyper-plane ($\vec{\psi}_j^\sour$). $C_1$ and $C_2$ are the coefficients which control the amount of similarity to the source-domain hyper-plane and the amount of classification correctness on the target domain, respectively. Once we have $\vec{\psi}_j^\adapt$, the new $\tau_j^\adapt$ is obtained similarly than in step 2.2) of \alg{trainNode}, {\ie} $\tau_j^\adapt = \argmax_{\tau} I(S_j^\adapt;\phi_j,\vec{\psi}_j^\adapt,\tau)$. Now, we have the new parameters defining the adapted root node, {\ie} $\theta_j^\adapt=(\phi_j,\vec{\psi}_j^\adapt,\tau_j^\adapt)$ ($j=1$). This means that we can apply $h(\vec{v}_k^\adapt;\theta_j^\adapt) \,\, \forall \vec{v}_k^\adapt \in S_j^\adapt$; therefore, we obtain the split $S_j^\adapt = S_j^{l,\adapt} \cup S_j^{r,\adapt}$. Next, we apply the same procedure for the child nodes of the root using $S_j^{l,\adapt}$ and $S_j^{r,\adapt}$ as input for the left and right nodes, respectively. The process continues until the leaves of the $i^{th}$ tree are reached. If the conditions for establishing a leaf node (the same than for learning source-domain trees) fulfill in a splitting node of the $i^{th}$ tree, then it becomes a leaf node and its corresponding posterior density is computed using the target-domain samples that reached it. Note that this can happen because the original structure of the $i^{th}$ tree was learned with $n$ source-domain samples, while the adaptation is based on the $m$ target-domain samples, and we assume $m \ll n$. Thus, the adapted tree is a reshaped version of the original one (\fig{cloningTrees}). 
%Moreover, in this case, the covariate shift assumption is retained for each pair of corresponding source-domain and adapted nodes of each tree $i$. For instance, for the path highlighted in \fig{cloningTrees} we have:
In addition, since we apply A-SVM at node level, we need to show that the covariate shift assumption is valid between two corresponding nodes of the source and target domains, which we prove as follows:
\begin{equation} \label{eq:demoCSNodeAdapt}
\begin{array}{l}
P^\sour(y|\vec{v}) \simeq P^\tar(y|\vec{v})\\
\begin{array}{ll}
\Rightarrow & P_i^\sour(y|\phi_1(\vec{v})) \simeq P_i^\tar(y|\phi_1(\vec{v}))\\
\Rightarrow & [\vec{\psi^\sour_1}, \tau^\sour_1] \simeq [\vec{\psi^\tar_1}, \tau^\tar_1] \quad \mbox{(SVM and} \, I(S_1^\tar;\phi_1,\vec{\psi}_1^\tar,\tau) \,  \mbox{are related to the posteriors in the domains})\\ 
\Rightarrow &  P_i^\sour(y|\vec{\psi^\sour_1}.\phi_1(\vec{v})+\tau^\sour_1) \simeq P_i^\tar(y|\vec{\psi^\tar_1}.\phi_1(\vec{v})+\tau^\tar_1)\\
\Rightarrow &  P_i^\sour(y|\vec{\psi^\sour_1}.\phi_1(\vec{v})+\tau^\sour_1\geq 0) \simeq P_i^\tar(y|\vec{\psi^\tar_1}.\phi_1(\vec{v})+\tau^\tar_1\geq 0) \quad \mbox{(posteriors of the samples which}\\
&\mbox{reach to the second node in the source and target domains are similar)}\\
\Rightarrow &  [\vec{\psi^\sour_2}, \tau^\sour_2] \simeq [\vec{\psi^\tar_2}, \tau^\tar_2]\\
\Rightarrow &  ...\\
\Rightarrow &  P_i^\sour(y|\vec{\psi^\sour_1}.\phi_1(\vec{v})+\tau^\sour_1,...,\vec{\psi^\sour_d}.\phi_d(\vec{v})+\tau^\sour_d) \simeq P_i^\tar(y|\vec{\psi^\tar_1}.\phi_1(\vec{v})+\tau^\tar_1, ..., \vec{\psi^\tar_d}.\phi_d(\vec{v})+\tau^\tar_d)\\
\Rightarrow &  P_i^\sour(y|\vec{\psi^\sour_1}.\phi_1(\vec{v})+\tau^\sour_1\geq0,...,\vec{\psi^\sour}.\phi_d(\vec{v})+\tau^\sour_d\geq0) \simeq  \\
            &  P_i^\tar(y|\vec{\psi^\tar_1}.\phi_1(\vec{v})+\tau^\tar_1\geq0, ..., \vec{\psi^\tar_d}.\phi_d(\vec{v})+\tau^\tar_d\geq0) . \\
\end{array}
\end{array}
\end{equation}
Of course, this reasoning can be extended to all other paths in the tree.

\subsection{Path-Adapt}
\label{sec:path-adapt}
Path-Adapt performs adaptation for each path of each source-domain tree instead of for each node individually; {\ie} for all splitting nodes in a given path, the adaptation is done simultaneously. Therefore, we need to show that the covariate shift assumption holds at path level, {\ie} beyond node level as required for Node-Adapt. In order to do that, as for Node-Adapt, we assume that the structure of the source-domain forest, $F^\sour$, is similar to the target-domain forest we are searching for, $F^\tar$. Thus, according to \equ{demoCSNodeAdapt}, we can show the relation  $P_i^\sour(y|\vec{\psi^\sour_1}.\phi_1(\vec{v})+\tau^\sour_1,\vec{\psi^\sour_2}.\phi_2(\vec{v})+\tau^\sour_2,...,\vec{\psi^\sour_d}.\phi_d(\vec{v})+\tau^\sour_d) \simeq P_i^\tar(y|\vec{\psi^\tar_1}.\phi_1(\vec{v})+\tau^\tar_1, \vec{\psi^\tar_2}.\phi_1(\vec{v})+\tau^\tar_2, ..., \vec{\psi^\tar_d}.\phi_d(\vec{v})+\tau^\tar_d)$, is established between each path of the $i^{th}$ tree of $F^\tar$ and its corresponding path of the $i^{th}$ tree of $F^\sour$, where $d$ is the index of the last splitting node in the $p^{th}$ path of the $i^{th}$ tree. Now, we define $Q^\sour_p = [\vec{\psi^\sour_1}, \tau^\sour_1; \vec{\psi^\sour_2}, \tau^\sour_2; ...; \vec{\psi^\sour_d}, \tau^\sour_d]$ and $Q^\tar_p = [\vec{\psi^\tar_1}, \tau^\tar_1; \vec{\psi^\tar_2}, \tau^\tar_2; ...; \vec{\psi^\tar_d}, \tau^\tar_d]$ as two mapping functions for projecting the feature set $\Phi_p = [\phi_1(\vec{v}),1; \phi_2(\vec{v}),1; ...; \phi_d(\vec{v}),1]$ to a new feature space. Let $[\vec{W}_p^\sour, b^\sour_p]$ be the SVM hyper-plane that we could learn in the $Q^\sour_p\cdot\Phi_p(\vec{v})=(\vec{\psi^\sour_1}\cdot\phi_1(\vec{v})+\tau^\sour_1,\ldots,\vec{\psi^\sour_d}\cdot\phi_d(\vec{v})+\tau^\sour_d)$ $d$-dimensional feature space, using sufficient source-domain data. Let $[\vec{W}_p^\tar, b^\tar_p]$ the analogous case using $Q^\tar_p\cdot\Phi_p(\vec{v})$ and target-domain data. Then, $[\vec{W}_p^\sour, b^\sour_p]$  and $[\vec{W}_p^\tar, b^\tar_p]$ should be similar since, according to \equ{demoCSNodeAdapt}, the relations $Q^\sour_p \simeq Q^\tar_p$ and $P_i^\sour(y|Q^\sour_p\cdot\Phi_p(\vec{v})) \simeq P_i^\tar(y|Q^\tar_p\cdot\Phi_p(\vec{v}))$ exist for every path of the tree (i.e. the covariate shift assumption in path level).

Accordingly, the strategy of Path-Adapt starts by searching for a $\tilde{Q}^\tar_p$ function so that, after projecting the target-domain samples in $S^\adapt$ to the new feature space, they can be class-discriminated successfully by $[\vec{W}_p^\sour, b^\sour_p]$. In particular, using the same structure than $F^\sour$ (same $T$, maximum depth of each path, and $\phi_j(\vec{v})$ feature selectors), as well as $S^\adapt$, a new forest $\tilde{F}^\tar$ is learned. Note that in this process there is no randomness, $\tilde{F}^\tar$ is a reshaped version of $F^\sour$ (leaves change, remind $m \ll n$). Now, we can define $\tilde{Q}^\tar_p=[\tilde{\psi}^\tar_1,\tilde{\tau}^\tar_1; \tilde{\psi}^\tar_2, \tilde{\tau}^\tar_2; ...; \tilde{\psi}^\tar_d, \tilde{\tau}^\tar_d]$ for the $p^{th}$ path of $\tilde{F}^\tar$. At this point, Path-Adapt uses $S^\adapt$ and $[\vec{W}_p^\sour, b^\sour_p]$ to adapt $\tilde{Q}^\tar_p$. However, the dimensionality of $\tilde{Q}^\tar_p$ can be eventually too high depending on the length of the path and the particular feature selectors in it, while at the same time the number of samples in $S^\adapt$ that eventually are considered along the path can be too low. Therefore, we propose to adapt only the thresholds. In particular, we propose the following optimization problem:
\begin{equation} \label{eq:node-adapt}
%\begin{array}{l}
\begin{split}
&B^{t_{a}}= argmin_{B^t} \frac{1}{2}||B^t-\tilde B^t||^2+C\sum_{p=1}^{R_i}\sum_{k=1}^{m}\epsilon_{k,p};\\
&s.t:\,\, \epsilon_{k,p}\geq0 \wedge y_k(\vec{W}_p^\sour\cdot(\tilde{Q}^\tar_p\cdot\Phi_p(\vec{v}_k^\adapt)+b^\sour_p))\geq -\epsilon_{k,p}\,\,\forall k\in\left\{1,\ldots,m\right\},\, p\in\left\{1,\ldots,R_i\right\}, \\
&\\
\end{split}
%\end{array}
\end{equation}
\noindent where $B^t =[\tau^t_1, \tau^t_2, ...,\tau^t_N]$, and $\tilde{B^t}=[\tilde{\tau}^t_1, \tilde{\tau}^t_2, ...,\tilde{\tau}^t_N]$. $N$ is equal to the number of nodes and $R_i$ stands for the number of paths in the $i^{th}$ tree. $C$ is a coefficient that controls the amount of similarity to the source model. We also restrict the adapted thresholds ($B^{t_a}$) to be similar to the target thresholds ($\tilde B^t$) to make balance between similarity to the source model and correctness of classification on target domain. \equ{node-adapt} is a QP convex optimization problem that can be solved with standard optimization packages. Finally, after obtaining the new thresholds for each tree, $S^\adapt$ is used again for reshaping and computing the posterior probabilities at the new leaves, so obtaining a final adapted forest $F^\adapt$. 

The hyper-plane $[\vec{W}_p^\sour, b^\sour_p]$ is computed after the training of $F^\sour$. In other words, for further domain adaptation, Path-Adapt requires to learn the different $[\vec{W}_p^\sour, b^\sour_p]$ as source-domain work. In fact, for each tree of $F^\sour$, and for each path in a given tree, we need to learn and store different $[\vec{W}_p^\sour, b^\sour_p]$ since we cannot know a priori which splitting nodes in $F^\sour$ will become leaves in $\tilde{F}^\tar$ and, therefore, which will be the final depth of a given path $p$. In particular, if $d$ is the length a path of a tree in $F^\sour$, then it is required to learn $d$ linear SVMs for such a path, using the source-domain samples. However, this is not necessarily quite time consuming since we even expect $dim(Q^s_p.\Phi_p) \ll  dim(\phi_j(\vec{v}))$ for each node $j$ in the path. Moreover, note that during Path-Adapt, the $\vec{\psi}_j^\sour$ are not used. Intuitively, such a collection of hyper-planes acts as a compact proxy of the source-domain data and the local experts $\vec{\psi}_j^\sour$ during the Path-Adapt process.

\subsection{Tree-Adapt}
\label{sec:tree-adapt}

According to the covariate shift assumption between source and target domains, in Tree-Adapt we replace a ratio $C$ of randomly selected trees in $F^\sour$, by an equivalent number of trees trained with $S^\adapt$ from the scratch. If we name as $F^\adapt_m$ the forest formed by such target-oriented set of trees, then joining $F^\adapt_m$ and $F^\sour$, gives rise to the adapted forest $F^\adapt$ of $T$ trees. Thus, $F^\adapt$ has a posterior distribution based on $F^\sour$ and $F^\adapt_m$, balanced by $C$:
\begin{equation} \label{eq:tree-adapt}
P(y|\vec{v})=\frac{1}{T}\left(\sum_{i=1}^{CT} P_i^\adapt(y|\vec{v})+\sum_{i=1}^{(1-C)T}P_i^\sour(y|\vec{v})\right), \mbox{where} \quad 0<C\leq1 .
\end{equation}

\section{Experimental results}
\label{sec:experiments}

\begin{figure} 
    \begin{subfigure}{0.24\textwidth}
         \includegraphics[ height=3cm, width=3.1cm]{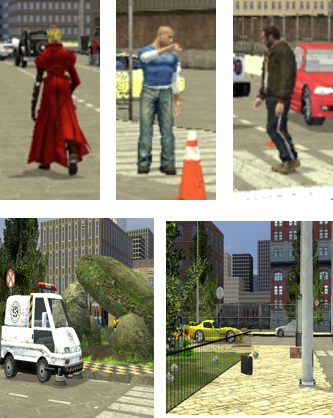}
        \caption{Virtual}
        \label{fig:v}
    \end{subfigure}
    \begin{subfigure}{0.24\textwidth}
         \includegraphics[ height=3cm, width=3.1cm]{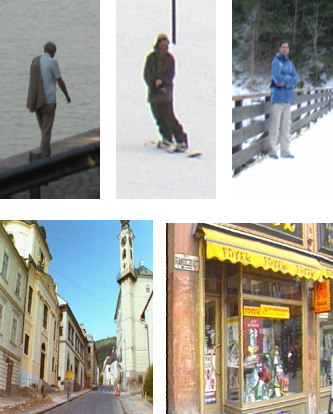}
        \caption{INRIA}
        \label{fig:I}
    \end{subfigure}
    \begin{subfigure}{0.24\textwidth}
        \includegraphics[ height=3cm, width=3.1cm]{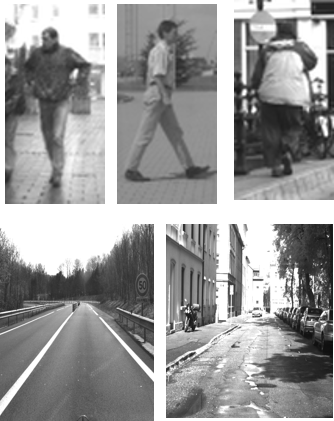}
        \caption{Daimler}
        \label{fig:D}
    \end{subfigure}    
    \begin{subfigure}{0.24\textwidth}
         \includegraphics[ height=3cm, width=3.1cm]{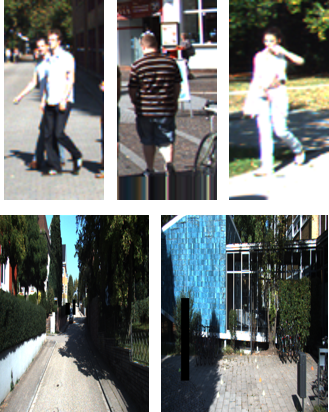}
        \caption{KITTI}
        \label{fig:K}
    \end{subfigure}    
    \caption{Pedestrians and backgrounds of four different datasets. The samples of these datasets show differences due to the camera (resolution, B/W-{\vs}-color, sensor sensitivity, focal length, lens distortion);  image acquisition conditions (usual viewpoint, scene illumination, backgrounds); and even the typical human poses and clothes.}\label{fig:datasets}
\end{figure}

In this section we show the effectiveness of the proposed DA methods for RF-LE. We present quantitative results on vision-based pedestrian detection as a challenging proof-of-concept object detection problem. We select four widely used pedestrian datasets, namely Virtual \cite{Vazquez:2014}, INRIA \cite{Dalal:2005}, Daimler \cite{Enzweiler:2009} and KITTI \cite{Geiger:2012}, to evaluate the RF-DA methods. In the case of KITTI we follow the train/test partition proposed in \cite{Premebida:2014}. Pedestrians and backgrounds from these datasets can be seen in \fig{datasets}. 

\subsection{Experimental setup \& implementation details}
The Virtual dataset is used only as source domain. The rest are considered both as source and target domains thanks to the use of different training and testing sets per dataset. % (\tab{datasets}). 
In order to run a DA experiment we select a relatively low number of the samples available in the training set of the target-domain dataset. For INRIA and KITTI this is $\sim 10\%$, while for Daimler we use $\sim 5\%$ since it is a larger training dataset.

Regarding the training of the RF-LE, we follow the instructions reported in \cite{Marin:2013}, {\eg} we use 100 trees and the maximum depth for the trees is set to 7, and HOG+LBP features are used as image descriptors.

The average miss rate (AMR) vs false positives per image (FPPI) \cite{Dollar:2012} to report detection accuracy. To set the hyper-parameters of the different adaptation methods ({\ie} the $C$'s), we used the Virtual-INRIA DA problem. These values are then used for the rest of experiments. 
%In particular, for Node-Adapt we use $C_1=1$ and $C_2=0.1$, for Path-Adapt $C_1=0.5\times10^{-5}$, and for Tree-Adapt $C=0.5$, respectively.

%\begin{table}[t]
%  \caption{Pedestrians and background samples in each dataset (Train/Test).}
%  \label{tab:datasets}
%  \centering
%  \begin{tabular}{lllll} \toprule
%                  & Virtual    & INRIA       & Daimler      & KITTI      \\ \midrule
%    Pedestrians   & 2,000 / 0  & 1,208 / 563 & 1,208 / 1193 & /  \\
%    Backgrounds   & 1,417 / 0  & 1,218 / 453 & 1,218 / 976  & /  \\
%    \bottomrule
%  \end{tabular}
%\end{table}

\begin{table}[t]
  \caption{The average miss rate (AMR) for 11 points in the interval of [0.01, 1] FPPI is calculated and shown for different baselines (Src, Tar100\%, TarX\% where X = 10 for INRIA and KITTI and X=5 for Daimler) and RF-DA methods (Node-Adapt, Path-Adapt and Tree-Adapt). Where V = Virtual, I = INRIA, D = Daimler, K = KITTI. Each "A $\rightarrow$ B" experiment is run five times to provide a mean AMR and corresponding standard deviation; thus the lower the better. Note how, Node-Adapt, Path-Adapt and Tree-Adapt show successful adaptation in most combinations (comparing with Src and TarX\%, even with Tar100).}
  \label{tab:Results1}
  \centering
  \begin{tabular}{lllllll}\toprule
                        &Src              &Tar100$\%$       &Tar X$\%$        &Node-Adapt       &Path-Adapt                  &Tree-Adapt \\\midrule
    V $\rightarrow$ I 	&19.06$\pm$1.77   &11.09$\pm$1.12  	&17.74$\pm$2.17   &16.37$\pm$1.84   &16.70$\pm$1.47   &14.59$\pm$1.39\\
    V $\rightarrow$ D   &31.95$\pm$1.35   &25.57$\pm$1.36  	&30.98$\pm$3.06   &27.95$\pm$3.70   &29.41$\pm$2.70     &25.39$\pm$1.94		\\
    V $\rightarrow$ K   &51.07$\pm$2.40   &36.02$\pm$2.28  	&53.06$\pm$3.93   &45.55$\pm$3.98   &38.95$\pm$2.36    &41.59$\pm$2.20		\\
    I $\rightarrow$ D   &23.28$\pm$1.65   &25.57$\pm$1.36  	&30.77$\pm$3.08   &26.72$\pm$3.69   &26.69$\pm$4.46    &22.87$\pm$2.30		\\
    I $\rightarrow$ K   &44.41$\pm$1.79   &36.02$\pm$2.28  	&53.06$\pm$3.93   &44.88$\pm$3.07   &38.84$\pm$4.37    &37.24$\pm$1.95		\\
    D $\rightarrow$ I   &21.98$\pm$1.29   &11.09$\pm$1.12  	&17.74$\pm$2.17   &14.77$\pm$1.58   &14.66$\pm$2.27    &13.92$\pm$1.17		\\
    D $\rightarrow$ K   &58.96$\pm$1.85   &36.02$\pm$2.28  	&53.06$\pm$3.93   &44.91$\pm$2.42   &41.58$\pm$2.97    &42.10$\pm$2.25		\\
    K $\rightarrow$ I   &48.19$\pm$1.57   &11.09$\pm$1.12  	&17.74$\pm$2.17   &17.74$\pm$3.21   &16.22$\pm$2.19    &18.17$\pm$1.92		\\
    K $\rightarrow$ D   &44.87$\pm$2.37   &25.57$\pm$1.36  	&30.77$\pm$3.08   &26.41$\pm$3.43   &27.41$\pm$2.81    &26.36$\pm$2.10		\\ \bottomrule
  \end{tabular}
\end{table}

\subsection{Results}
In order to show the effectiveness of the proposed RF-DA methods, we run experiments for different combinations of source and target domains. Each train-test experiment is repeated five times and we report the mean and standard deviation of AMR versus FPPI in the interval of [0.01, 1]. We compare Node-Adapt, Path-Adapt and Tree-Adapt methods in each experiment with the RF trained only with source data (Src), the RF trained with all samples of the target domain (Tar100\%), and with the RF trained only on the limited number of target-domain samples(TarX\%). 

In this case, DA is successful if reduces the AMR of both Src and TarX\%, and is close to the AMR of Tar100\%. Therefore, we can see in \tab{Results1}, that Node-Adapt, Path-Adapt and Tree-Adapt provide successful results. Notably, Tree-Adapt, the simplest strategy, is more successful than the other two. 

Node-Adapt and Path-Adapt report similar performance. However, when the number of available samples in target domain is limited, the probability of error propagation becomes higher for Node-Adapt (too few samples to train the local experts) and, a priori, Path-Adapt should be more successful since it uses all samples for adapting to adapt less parameters. Tree-Adapt can suffer the same problem that Node-Adapt since the learned target-domain trees can be too shallow. To assess this point, we repeat the same experiments with half of available target-domain samples. \tab{lesssample-table} shows the corresponding results\footnote{The information in \tab{Results1} and in \tab{lesssample-table} is plotted as AMR-{\vs}-FPPI curves in the supplementary material.}. 

Indeed, Path-Adapt is more effective than Node-Adapt in this demanding scenario. However, Tree-Adapt still is the most effective method. On the other hand, for low memory adaptive embedded systems ({\eg} a domestic cleaning robot), Path-Adapt is interesting in the sense that it requires scarce information from the source domain since the data and node hyper-planes are not required, just the source tree topology and the compact path SVMs are needed. So the system can come with a sort of common knowledge and be (self-)adapted for the particular environment. After all, Path-Adapt results are just scarcely worse than those of Tree-Adapt. 
  
\begin{table}
  \caption{Analogous to \tab{Results1} but for a more challenging situation in terms of available target-domain data. In particular, X = 5 for INRIA and KITTI and X = 3 for Daimler.}
  \label{tab:lesssample-table}
  \centering
  \begin{tabular}{lllllll}
    \toprule
          &Src      &Tar100$\%$   &Tar X$\%$   &Node-Adapt     &Path-Adapt &Tree-Adapt \\
    \midrule
    V $\rightarrow$ I 	&19.06$\pm$1.77   &11.09$\pm$1.12  	&24.97$\pm$3.85 	 &19.29$\pm$1.38   &18.77$\pm$2.72 &16.65$\pm$2.15\\
    V $\rightarrow$ D     &31.95$\pm$1.35   &25.57$\pm$1.36  	 &31.01$\pm$3.63 	&28.55$\pm$3.14   &29.56$\pm$4.80 &25.77$\pm$1.97		\\
    V $\rightarrow$ K     &51.07$\pm$2.40   &36.02$\pm$2.28  	&57.96$\pm$3.69 	 &44.24$\pm$4.26   &38.34$\pm$2.03 &43.36$\pm$2.28		\\
    I $\rightarrow$ D     &23.28$\pm$1.65   &25.57$\pm$1.36  	&31.71$\pm$3.23 	 &27.12$\pm$3.34   &28.65$\pm$3.78 &23.71$\pm$1.79		\\
    I $\rightarrow$ K     &44.41$\pm$1.79   &36.02$\pm$2.28  	&58.34$\pm$2.89 	 &44.33$\pm$2.41   &40.52$\pm$2.23 &39.79$\pm$2.87		\\
    D $\rightarrow$ I     &21.98$\pm$1.29   &11.09$\pm$1.12  	&20.62$\pm$5.53      &17.66$\pm$2.32 	 &16.33$\pm$2.73   &15.23$\pm$2.29 	\\
    D $\rightarrow$ K     &58.96$\pm$1.85   &36.02$\pm$2.28  	&56.41$\pm$2.35 	 &47.39$\pm$2.56   &45.48$\pm$2.75 &45.20$\pm$2.50		\\
    K $\rightarrow$ I     &48.19$\pm$1.57   &11.09$\pm$1.12  	&24.34$\pm$5.41 	 &21.02$\pm$2.84   &19.08$\pm$3.43 &19.31$\pm$3.18		\\
    K $\rightarrow$ D     &44.87$\pm$2.37   &25.57$\pm$1.36      &30.92$\pm$3.23  	&27.53$\pm$2.97 	&27.81$\pm$3.16   &26.36$\pm$3.44 \\
    \bottomrule
  \end{tabular}
\end{table}

Overall, we conclude that among the three tested strategies, Tree-Adapt is the simplest to implement and the one that offers better results. Interestingly, \tab{Results1} shows that the pedestrian detector based on the RF-LE trained with synthetic data (no manual labeling required) and adapted with a relatively low number of real-world labels using Tree-Adapt, performs not too far than an equivalent detector based on the labeling of 10-20 times more pedestrians (Tar100\%). In addition, as we have mentioned, Path-Adapt can be a good alternative depending on the application.

\section{Conclusions}
\label{sec:conclusions}   
We have proposed three novel supervised model-transfer DA methods for RF-LE models, namely Node-Adapt, Path-Adapt and Tree-Adapt. These methods are inherently different since they perform the adaption at different levels of the RF-LE. We have investigated the validity of the covariate shift assumption for them to support their approaches of adaptation. We have assessed their performance by facing the challenge of vision-based pedestrian detection. In particular, we have considered four well-known benchmark datasets as different domains. The conclusion is that, although these three proposals achieve DA, Tree-Adapt is the most successful method to adapt RF-LE models, and opens the door to the use of virtual worlds as source domain for reducing the number of object annotations. As future work we plan several working lines. First, we want to assess the possibility of combining the core ideas behind the proposed methods so that adaptation is further improved; alternatively they can be used to self-annotate training data for progressively improving its accuracy (each method acting as a different oracle). Second, we want to assess how Tree-Adapt can be used in an online learning setting.

\newpage

\begin{footnotesize}

\bibliography{BibStringsShort,RFDA} 
\bibliographystyle{ieeetr}

\end{footnotesize}

\newpage
\section*{Supplementary Materials}
\textbf{1.Results of Table 1 as plots}

\begin{longtable}{cc}
\includegraphics[ height=5.7cm, width=7cm]{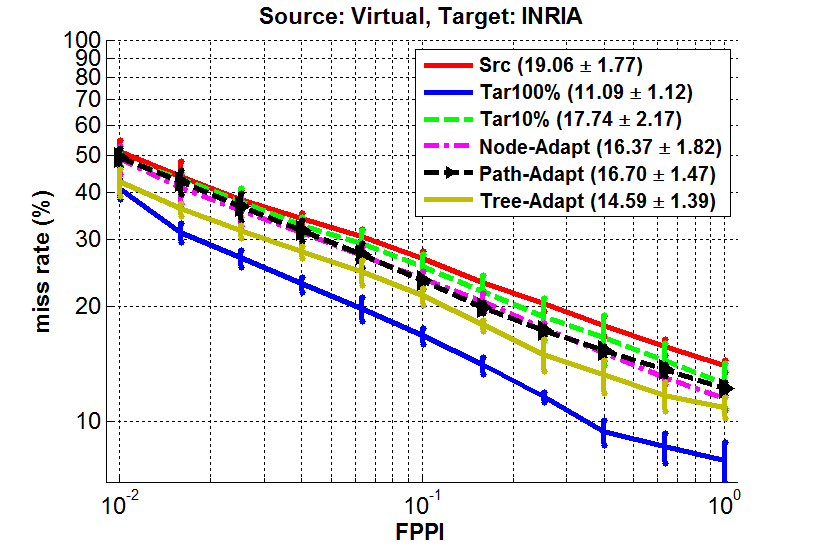} &
\includegraphics[ height=5.7cm, width=7cm]{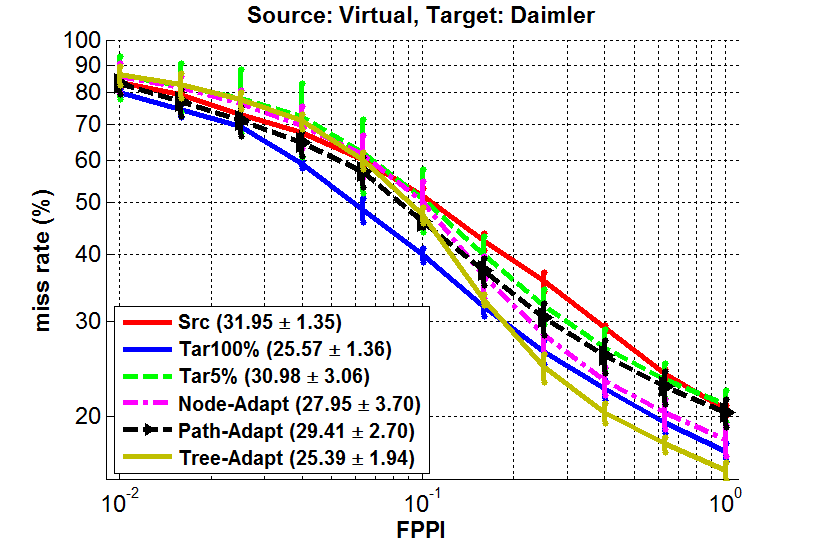} \\
(a) & (b) \\
\includegraphics[ height=5.7cm, width=7cm]{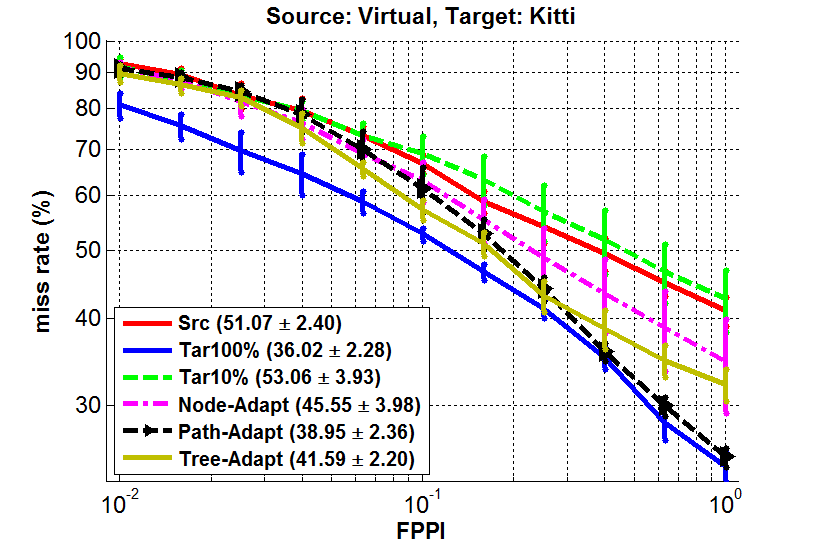} &
\includegraphics[ height=5.7cm, width=7cm]{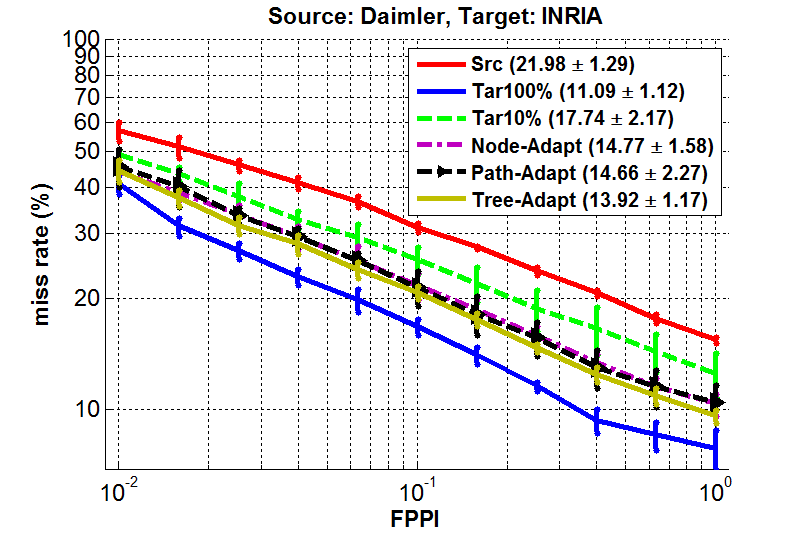} \\
(c) & (d) \\
\includegraphics[ height=5.7cm, width=7cm]{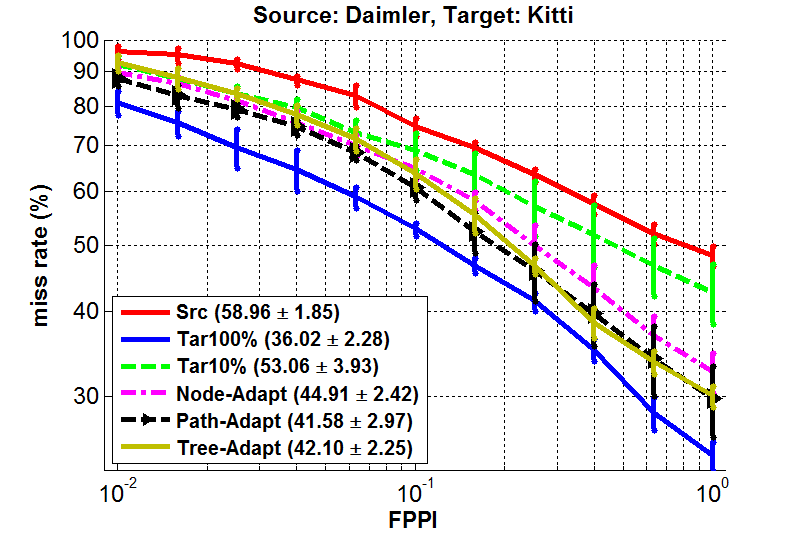} &
\includegraphics[ height=5.7cm, width=7cm]{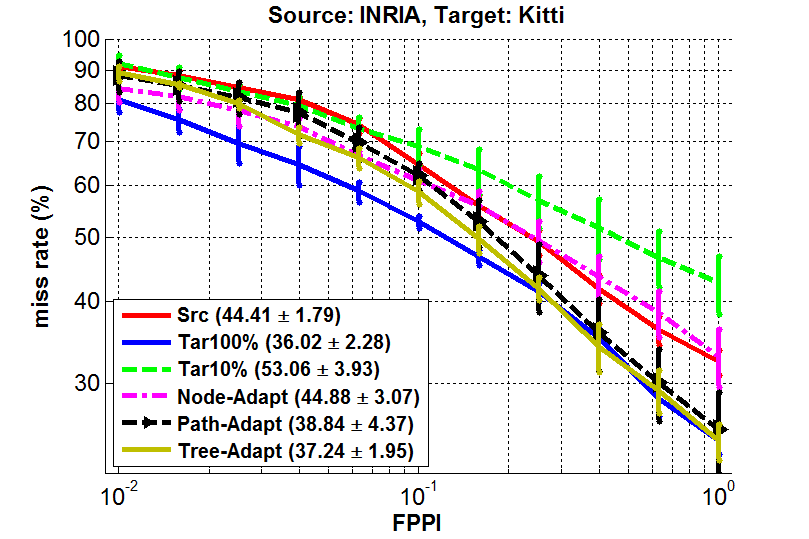} \\
(e) & (f) \\
 \includegraphics[ height=5.7cm, width=7cm]{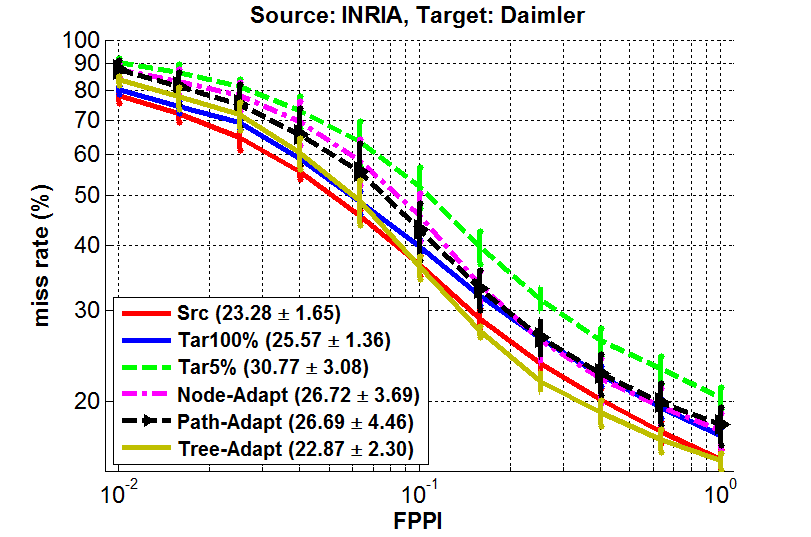} &
  \includegraphics[ height=5.7cm, width=7cm]{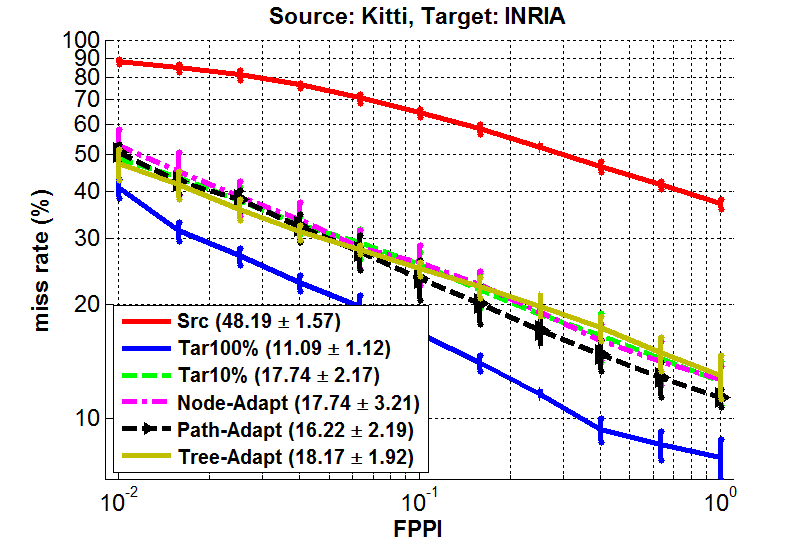} \\
 (g) & (h)\\
 \includegraphics[ height=5.7cm, width=7cm]{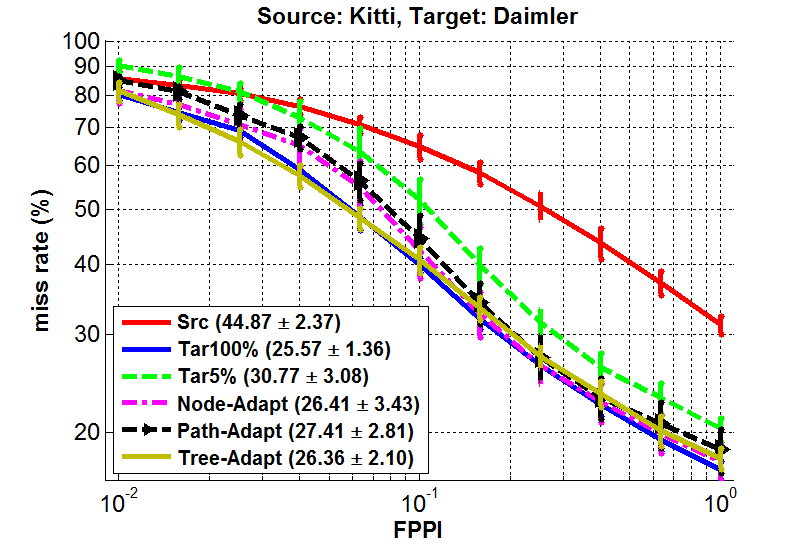} &\\
 (i) &
 
\end{longtable} 
\begin{figure} [h]
\caption{Average miss rate vs FPPI curve in the interval of [0.01, 1] is plotted for Node-Adapt, Path-Adapt and Tree-Adapt in comparison to different baselines: Src, Tar100\%, Tar10\% (when the target domains are Kitti or INRIA) and Tar5\% (when the target domain is Daimler). The experiment is repeated for all combination of the source and target domains. As can be seen, the proposed DA methods can achieve lower miss rate than Src and Tar10\% or Tar5\% baselines which indicates successful adaptation. Tree-Adapt is the most successful one. }
\end{figure}
\newpage
\textbf{2.Results of Table 2 as plots}

\begin{longtable}{cc}
\includegraphics[ height=5.7cm, width=7cm]{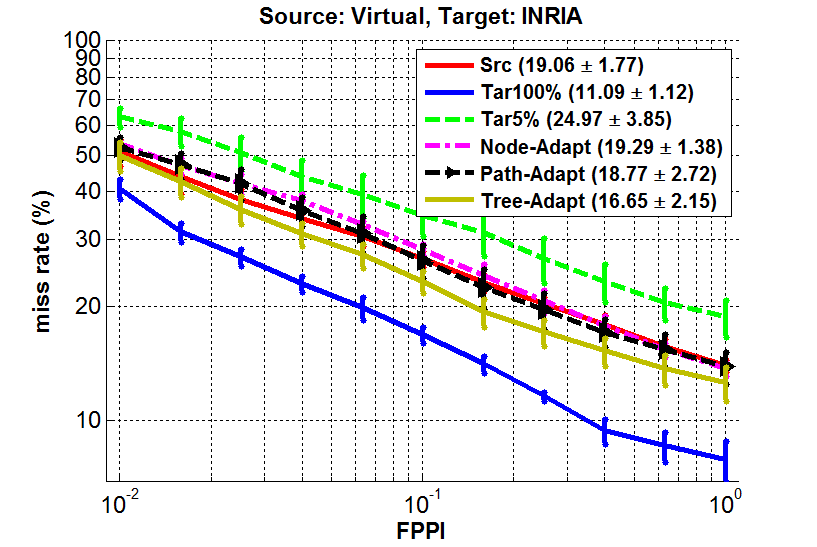} &
\includegraphics[ height=5.7cm, width=7cm]{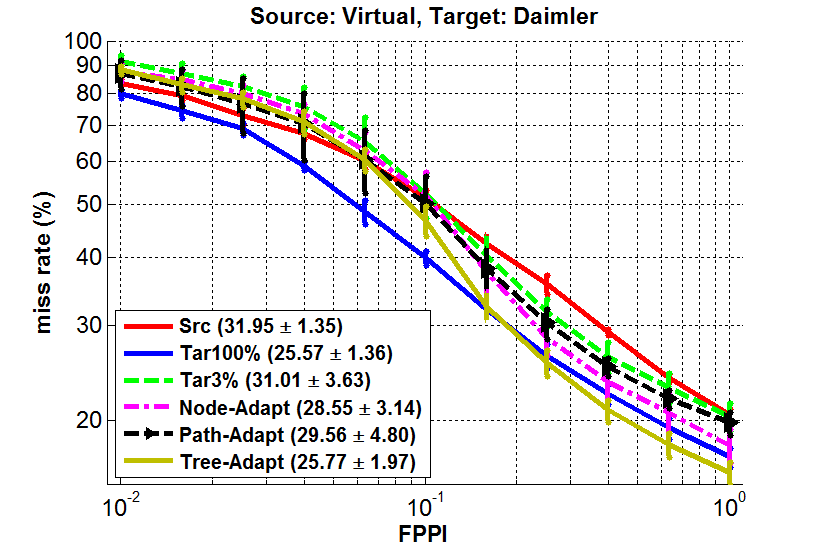} \\
(a) & (b) \\
\includegraphics[ height=5.7cm, width=7cm]{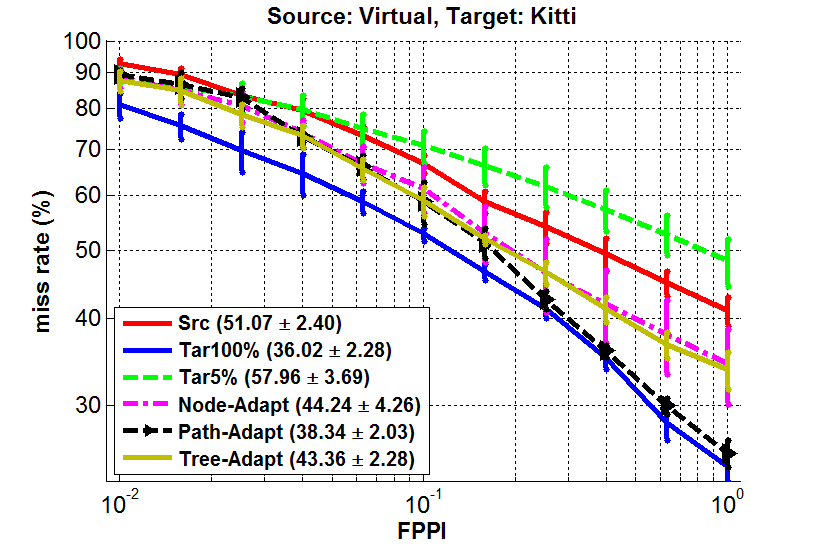} &
\includegraphics[ height=5.7cm, width=7cm]{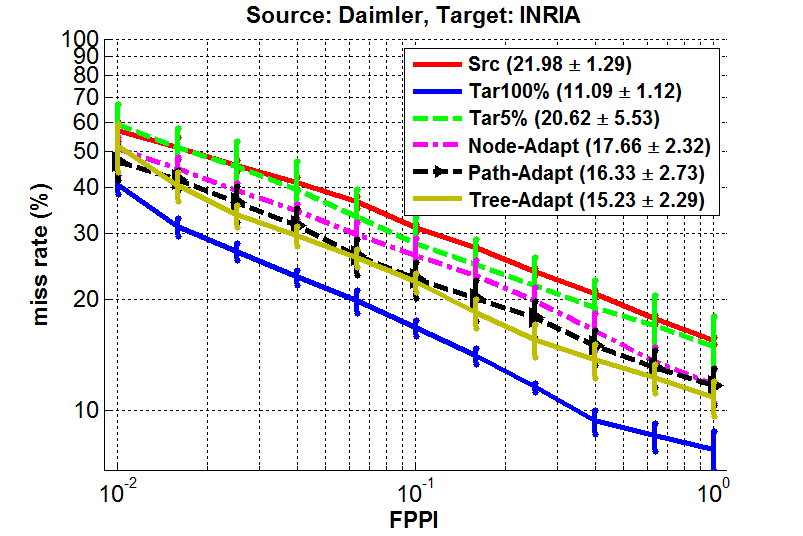} \\
(c) & (d) \\
\includegraphics[ height=5.7cm, width=7cm]{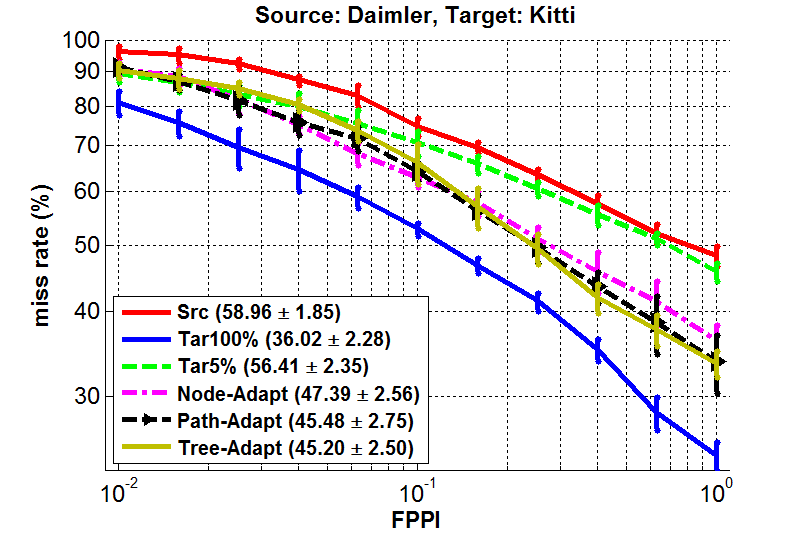} &
\includegraphics[ height=5.7cm, width=7cm]{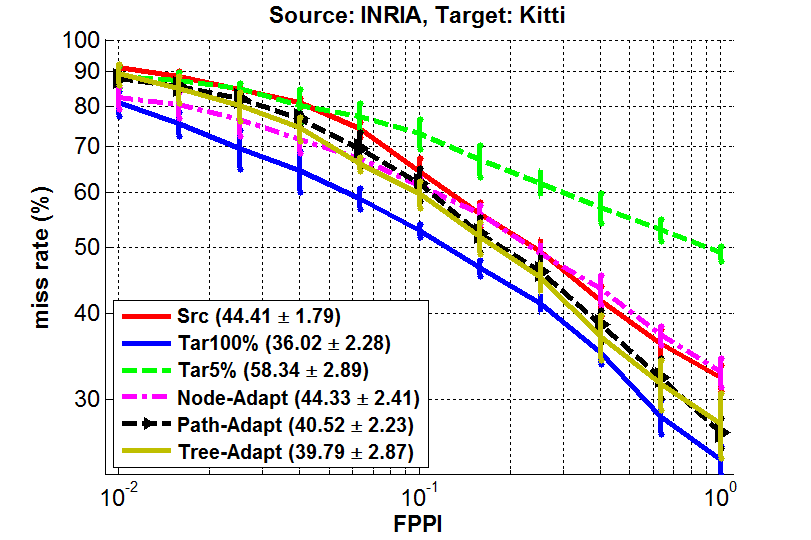} \\
(e) & (f) \\
 \includegraphics[ height=5.7cm, width=7cm]{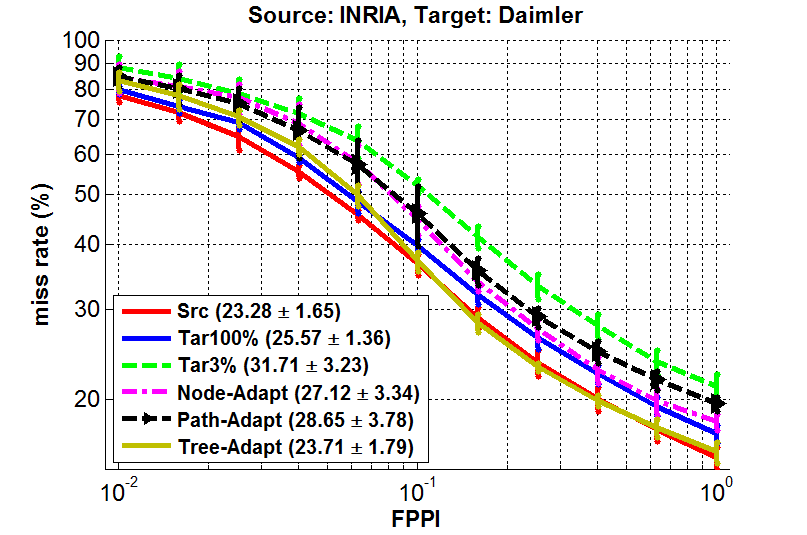} &
  \includegraphics[ height=5.7cm, width=7cm]{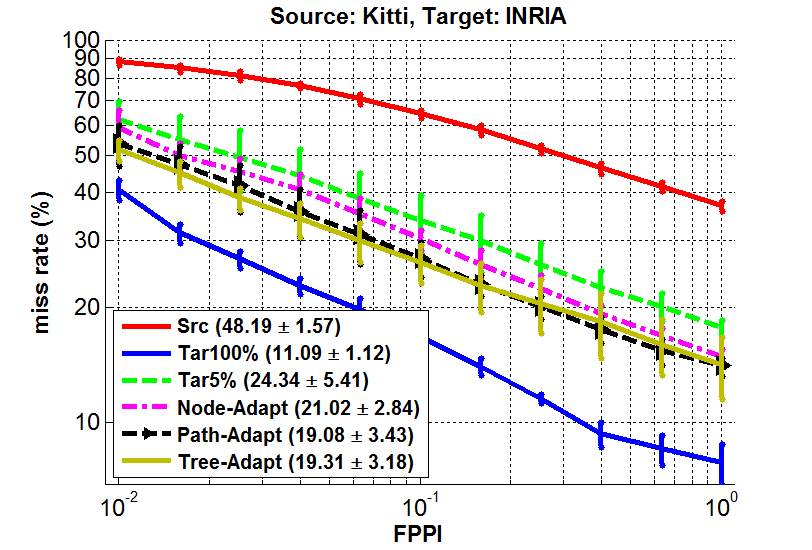} \\
 (g) & (h)\\
 \includegraphics[ height=5.7cm, width=7cm]{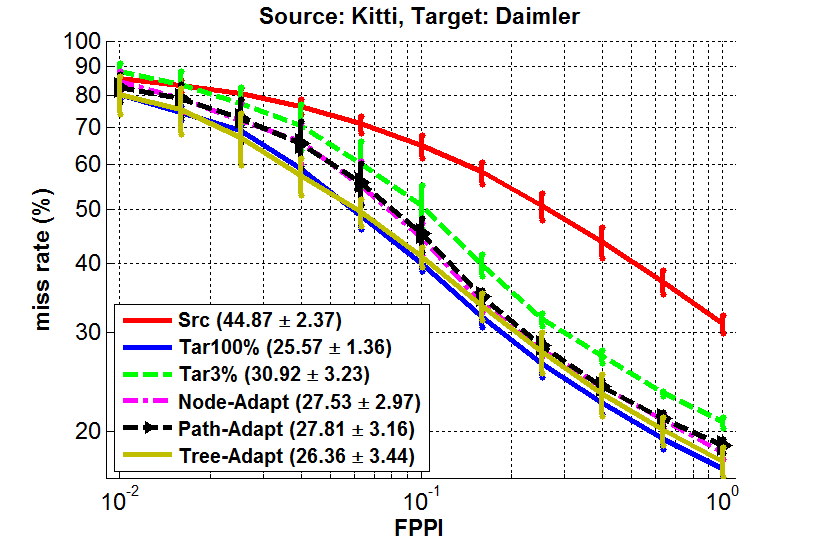} &\\
 (i) &
 
\end{longtable} 
\begin{figure} [h]
\caption{Average miss rate vs FPPI curve in the interval of [0.01, 1] is plotted for Node-Adapt, Path-Adapt and Tree-Adapt in comparison to different baselines: Src, Tar100\%, Tar5\% (for Kitti or INRIA as the target domain) and Tar3\% (for Daimler as the target domain) when the number of available target samples is very low. Comparing to Figure1, Node-Adapt obtains higher miss rate than Path-Adapt which indicates that the robustness of Path-Adapt is more than Node-Adapt in the few training target samples situation. }
\end{figure}

\end{document}